\documentclass[lettersize,journal]{IEEEtran}
\usepackage{amsmath,amsfonts}
\usepackage{algorithmic}
\usepackage{algorithm}
\usepackage{array}
\usepackage[caption=false,font=normalsize,labelfont=sf,textfont=sf]{subfig}
\usepackage{textcomp}
\usepackage{stfloats}
\usepackage{url}
\usepackage{verbatim}
\usepackage{graphicx}
\usepackage{cite}
\usepackage{color}
\begin{document}

\title{AdaptivePose++: A Powerful Single-Stage Network for Multi-Person Pose Regression}

\author{Yabo Xiao, Xiaojuan Wang$^*$, Dongdong Yu, Kai Su, Lei Jin, Mei Song, Shuicheng Yan, ~\IEEEmembership{Fellow,~IEEE,} Jian Zhao$^*$, ~\IEEEmembership{Member,~IEEE}

\thanks{Yabo Xiao, Xiaojuan Wang, Lei Jin, Mei Song are with School of Electronic Engineering, Beijing University of Posts and Telecommunications, Beijing, China.Email: \{xiaoyabo, wj2718, songm\}@bupt.edu.cn}

\thanks{Dongdong Yu is with OPPO Research Institute, Beijing, China. Email: yudongdong@oppo.com}

\thanks{kai Su is with ByteDance Inc., Beijing, China. Email: sukai@bytedance.com}


\thanks{Shuicheng Yan is with Sea AI Lab (SAIL), Singapore. Email: yansc@sea.com}

\thanks{ Jian Zhao is with Institute of North Electronic Equipment, Beijing, China and Department of Mathematics and Theories, Peng Cheng Laboratory, Shenzhen, China. Email: zhaojian90@u.nus.edu}

\thanks{Corresponding authors: Xiaojuan Wang and Jian Zhao}} 



\maketitle

\begin{abstract}
Multi-person pose estimation generally follows top-down and bottom-up paradigms. Both of them use an extra stage ($\boldsymbol{e.g.,}$ human detection in top-down paradigm or grouping process in bottom-up paradigm) to build the relationship between the human instance and corresponding keypoints, thus leading to the high computation cost and redundant two-stage pipeline. To address the above issue, we propose to represent the human parts as adaptive points and introduce a fine-grained body representation method. The novel body representation is able to sufficiently encode the diverse pose information and effectively model the relationship between the human instance and corresponding keypoints in a single-forward pass. With the proposed body representation, we further deliver a compact single-stage multi-person pose regression network, termed as AdaptivePose. During inference, our proposed network only needs a single-step decode operation to form the multi-person pose without complex post-processes and refinements. We employ AdaptivePose for both 2D/3D multi-person pose estimation tasks to verify the effectiveness of AdaptivePose. Without any bells and whistles, we achieve the most competitive performance on MS COCO and CrowdPose in terms of accuracy and speed. Furthermore, the outstanding performance on MuCo-3DHP and MuPoTS-3D further demonstrates the effectiveness and generalizability on 3D scenes. Code is available at \textcolor{red}{ https://github.com/buptxyb666/AdaptivePose.}
\end{abstract}




\section{Introduction}
\IEEEPARstart{H}{uman} pose estimation (HPE) \cite{hourglass,spcnet,openpose,grmi} a is classical yet challenging task in computer vision communities. It aims to locate the person keypoints from the natural image. HPE always serves as the necessary step for high-level vision tasks such as action recognition \cite{stgcn, asgcn, csvt4, csvt5} and pose tracking \cite{simplebaseline,csvt2}, etc. Existing 2D/3D multi-person pose estimation methods can be categorized into top-down\cite{cpn,grmi, hrnet, rmpe, maskrcnn, bytepose, csvt3} and bottom-up \cite{ openpose, ae, pifpaf, higherhrnet, personlab, re49, lqr,csvt1, csvt8} paradigms. The top-down strategy divides this problem into human detection and single-person pose estimation, each detected human region is cropped and normalized to locate the single-person keypoints. It achieves the superior performance while suffers the large computation cost and low efficiency due to the additional human detector. The bottom-up strategy formulates this task as keypoint localization and grouping process. It firstly detects all person keypoints simultaneously on the full image instead of the cropped single-person regions and then assigns them to individuals. Although bottom-up methods are more efficient than top-down methods, the heuristic grouping process is still computationally complex, and always involves many hand-designed rules. 

\begin{figure}
\begin{center}
\includegraphics[width=0.9\columnwidth]{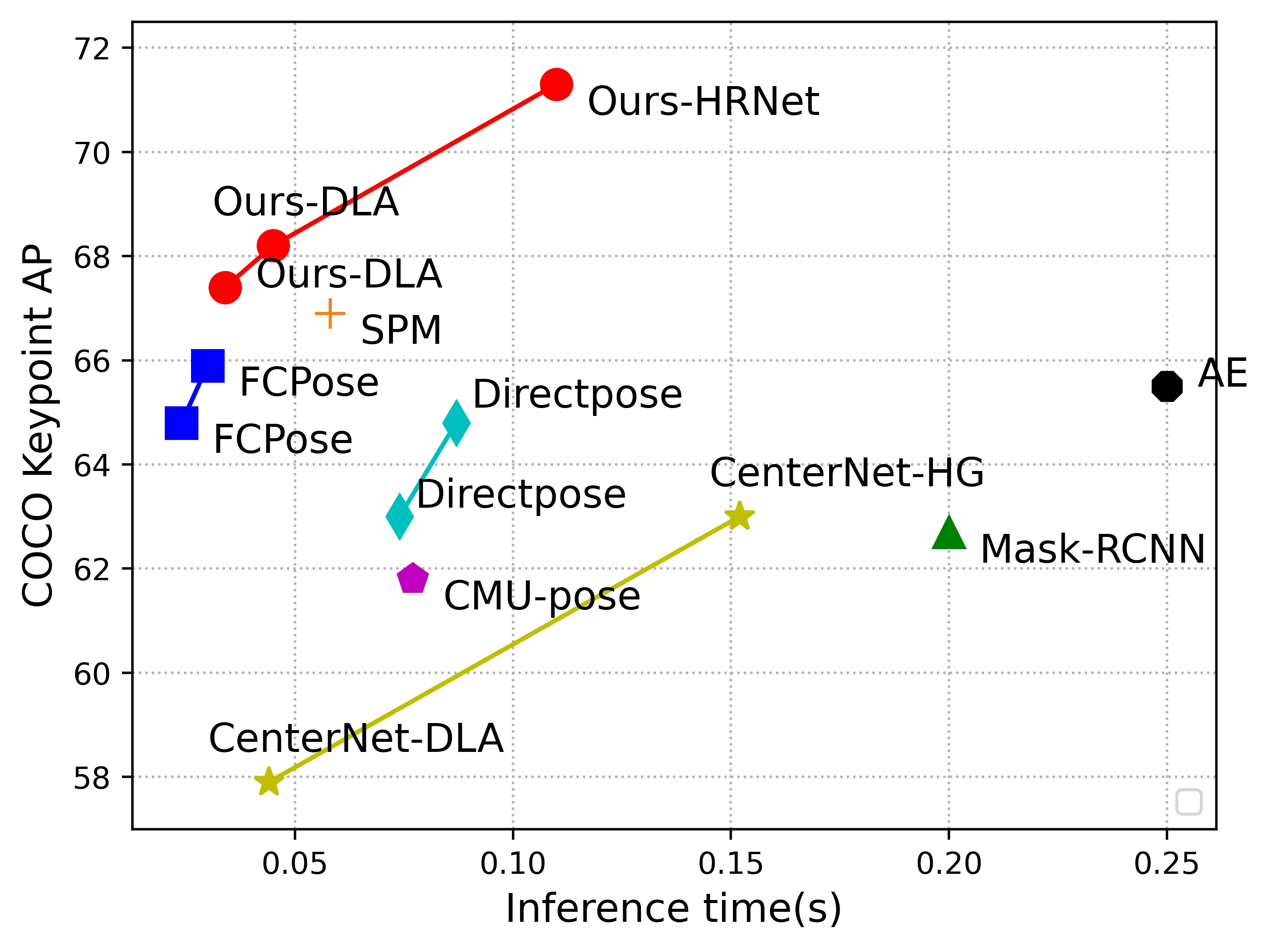}
\end{center}
\caption{Inference time~(s) $\boldsymbol{vs.}$ precision~(COCO keypoint AP). Our method achieves the best speed-accuracy trade-offs compared with previous methods on MS COCO \cite{coco}.}
\label{fig:speed-acc}
\end{figure}

\begin{figure}
\begin{center}
\includegraphics[width=1.0\columnwidth]{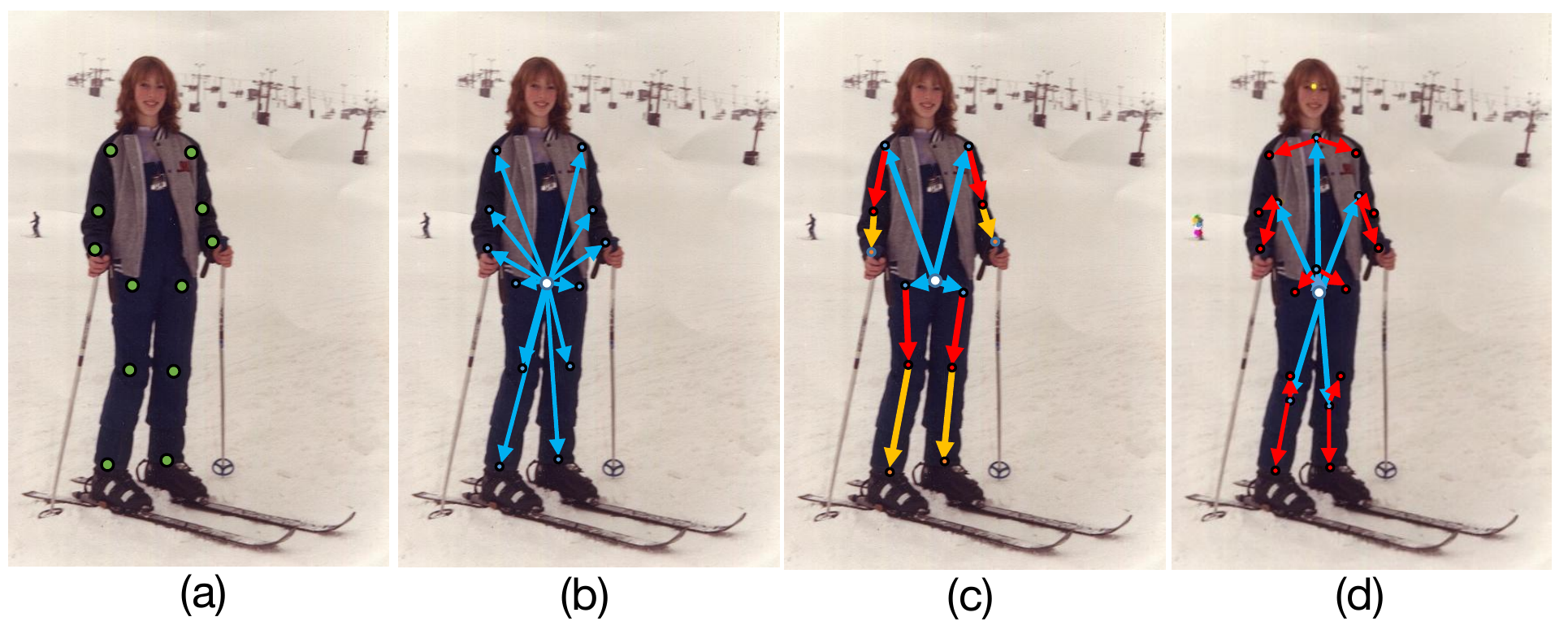}
\end{center}
\caption{(a) Conventional body representation generally used in top-down methods such as Rmpe \cite{rmpe} as well as bottom-up methods such as CMU-pose \cite{openpose}. (b) Center-to-joint body representation proposed by CenterNet \cite{centernet}. (c) Hierarchical body representation introduced by SPM \cite{spm}. (d) Our adaptive point set representation. In contrast to (b) and (c) only using center or root features, the features of adaptive points are introduced to encode the keypoint information in each part.}
\label{fig:bofy-rep}
\end{figure}

Both top-down and bottom-up methods generally use the conventional keypoint heatmap representation that models the human pose via absolute keypoint position, as shown in Fig. \ref{fig:bofy-rep} (a), which separates the relationship between the position of human instance and corresponding keypoints. Consequently, an extra stage is required to build up the connections. Recent research works have tried to model the connections between human body and corresponding keypoints in a single-forward process while suffering some obstacles, thus leading to the compromising performance. As shown in Fig. \ref{fig:bofy-rep} (b), CenterNet \cite{centernet} represents the instance as center point and encodes the relationship between instance and its keypoints via center-to-joint offsets. Nevertheless, it achieves inferior performance since the limited center feature can not encode the various pose effectively. As shown in Fig. \ref{fig:bofy-rep} (c), SPM \cite{spm} also represents the human instance via the limited feature of root joint and further employs a fixed hierarchical structure along the skeleton path to build the relationship between the human instance and keypoints. Due to the intermediate nodes are pre-defined and the supervision acting on the offsets between the adjacent joints, thus the fixed hierarchical path will lead to accumulated errors along the hierarchical path.



To address the aforementioned problems, in this work, we propose a novel body representation which is able to sufficiently encode various human pose and effectively build the relations between the instance and keypoints in a single-forward pass. Specifically, human body is divided into several parts and each human part is represented as an adaptive part related point. In this manner, we leverage the human center feature together with the features at several human-part related points to represent diverse human pose. The connections can be built by the center to adaptive points then to keypoints path as shown in Fig. \ref{fig:bofy-rep} (d). Compared with previous representations, our representation brings two-fold benefits as follows: 1)The proposed point set representation introduces additional features at adaptive part related points, which are able to encode more informative features for flexible pose compared with limited center representation. 2) The adaptive part related points serves as relay nodes can more effectively model the associations between human instance and corresponding keypoints in a single-forward pass.


With the adaptive point set representation, we propose an effective and efficient single-stage differentiable regression network, termed AdaptivePose, which mainly consists of three novel components. First, we introduce the Part Perception Module to regress seven adaptive human-part related points for perceiving corresponding seven human parts. Second, in contrast to using the limited feature with fixed receptive field to predict the human center, we propose the Enhanced Center-aware Branch to conduct the receptive field adaptation by aggregating the features of adaptive human-part related points to perceive the center of various pose more precisely. Third, we propose the Two-hop Regression Branch together with the Skeleton-Aware Regression Loss for regressing keypoints. The adaptive human-part related points act as one-hop nodes to factorize the center-to-joint offsets dynamically.

A preliminary version of this work \cite{adaptivepose} was accepted in AAAI
Conference on Artificial Intelligence (AAAI) 2022. We extend it in terms of five aspects: 1) We augment the content of the Abstract, Introduction, Related Work, Methodology and Experiments to cover sufficient details for clearer and more comprehensive presentation. 2) We improve the regression loss and add an additional loss term to learn the bone connections of inner parts and cross parts, which is helpful for crowd scene. 3) We tune several hyper-parameters and improve the performance in single forward pass, and add more ablation experiments with analyses to verify the superior positioning capacity of our framework. We further report the more comprehensive comparisons with competitive bottom-up counterparts and list more qualitative results. 4) We report the state-of-the-art results on the CrowdPose \cite{crowdpose}, which contains an enormous number of crowd scenes. 5) We keep the 2D framework and add the depth estimation components, further extent our method to 3D multi-person pose estimation task, the promising results on MuPoTS-3D \cite{muco} verify the effectiveness and generalizability of our method in 3D scenes.


We summarize our main contributions as follows:
\begin{itemize}
\item We propose to represent human parts as points thus the human body can be represented via an adaptive point set including center and several human-part related points. To our best knowledge, we are the first to present a fine-gained and adaptive body representation to sufficiently encode the pose information and effectively build up the relation between the human instance and keypoints in a single-forward pass.

\item Based on the novel representation, we exploit a compact single-stage differentiable network, termed as AdaptivePose. Specifically, we introduce a novel Part Perception Module to perceive the human parts by regressing seven human-part related points. By manipulating human-part related points, we further propose the Enhanced Center-aware Branch to more precisely perceive the human center and the Two-hop Regression Branch together with the Skeleton-Aware Regression Loss to precisely regress the keypoints.

\item Our method significantly simplifies the pipeline of existing multi-person pose estimation methods. The effectiveness is demonstrated on both 2D, 3D pose estimation benchmarks. We achieves the best speed-accuracy trade-offs without complex refinements and post-processes. Furthermore, extended experiments on CrowdPose and MuPoTS-3D clearly verify the generalizability on crowd and 3D scenes.

\end{itemize}

\section{Related work}
In this section, we review three parts related to our method including top-down methods, bottom-up methods and point-based methods.  

{\bf Top-down Methods.} Given an arbitrary RGB image, the top-down methods \cite{cpn,grmi, hrnet, rmpe, maskrcnn, bytepose, csvt3} crop and resize the region of detected person firstly and then locate the single-person keypoints in each cropped area. The detected human areas are cropped and resized to a unified size so that it has superior performance. For convolution-based methods, HRNet~\cite{hrnet} maintains high-resolution features and repeatedly fuses multi-resolution features through the whole process to generate reliable high-resolution representations. Su et al. \cite{bytepose} propose a Channel Shuffle Module and Spatial, Channel-wise Attention Residual Bottleneck~(SCARB) to drive the cross-channel information flow. Zhao et at.\cite{csvt3} leverage a quality prediction block (OKS-net) to regress object keypoint similarity, which builds the direct awareness of the predicted pose quality. For transformer-based network, TokenPose \cite{tokenpose} embeds each keypoint as a token to simultaneously learn constraint relationships across keypoints and visual representation from images. Other researches \cite{devil, crowdpose} try to handle quantization error and occlusion issue. However, the detection-first paradigm always brings additional computational cost and forward time, top-down methods are often not feasible for the real-time systems with strict latency constraints.

{\bf Bottom-up Methods.} In contrast to top-down methods, bottom-up methods \cite{ openpose, ae, pifpaf, higherhrnet, personlab, re49, lqr,csvt1, csvt8} first localize keypoints of all human instances in the input image and then group them to the corresponding person. Bottom-up methods mainly concentrate on the effective grouping process or tackling with the scale variation. For example, CMU-pose~\cite{openpose} proposes a non parametric representation, named Part Affinity Fields~(PAFs), which encodes the location and orientation of limbs, to group the keypoints to individuals. AE~\cite{ae} simultaneously outputs a keypoint heatmap and a tag map for each body joint, then assigns the keypoints with similar tags into individual. HigherHRNet~\cite{higherhrnet} generates high-resolution feature pyramid with multi-resolution supervision and multi-resolution heatmap aggregation for learning scale-aware representations. Li et al. \cite{csvt1} exploit an encoding-decoding network with multi-scale Gaussian heatmaps and guiding offset fields to represent multi-person pose information, and introduce an auxiliary task of peak regularization into heatmap supervision for improving performance. However, one case worth noting is that the grouping process serves as a post-process is still computationally complex and redundant. 

\begin{figure}
\begin{center}
\includegraphics[width=1.0\columnwidth]{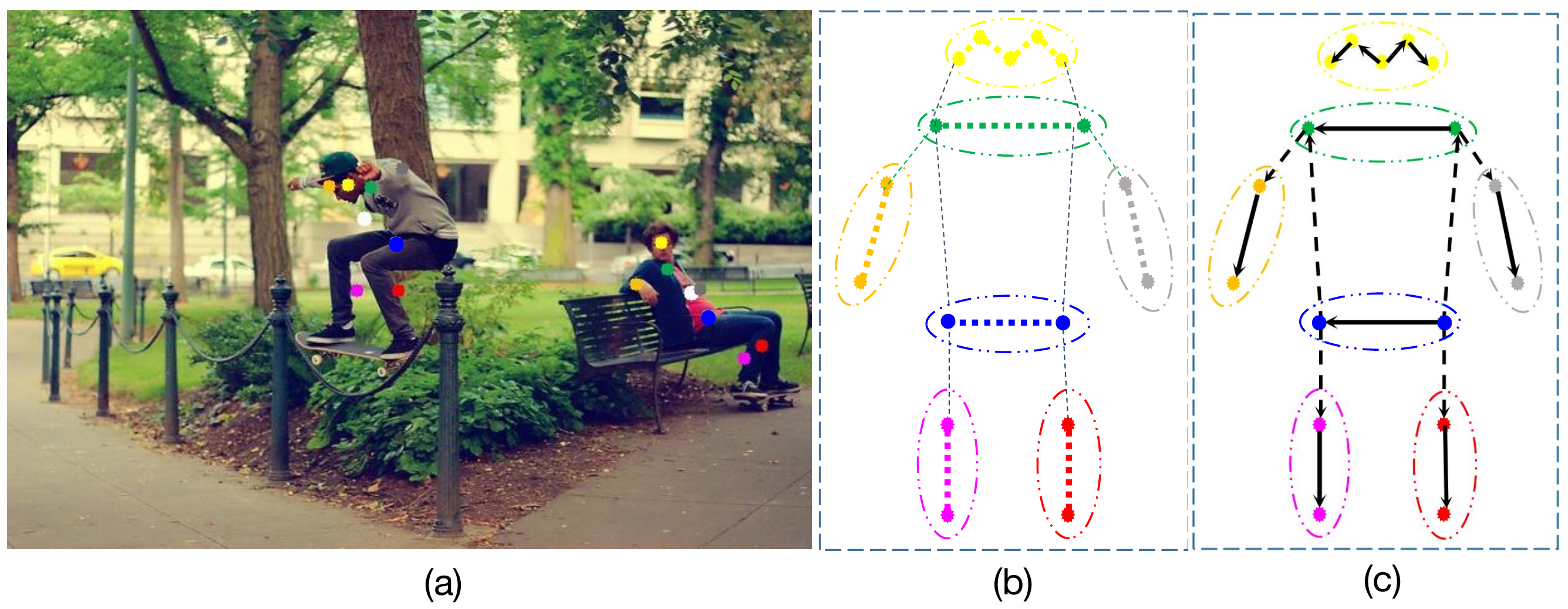}
\end{center}
\caption{(a) The visualization of adaptive point set. White points indicate the human center and others refer to part related points visualized by different colors. We leverage an adaptive point set conditioned on each human instance to represent the human pose in a fine-grained way. (b) Divided human parts according to inherent body structure. (c) Black dotted arrows indicate the bone connections of cross parts and solid lines refer to bone connections of inner parts.}
\label{fig:image3}
\end{figure}

{\bf Point-based methods.} Before the deep learning era, in contrast to most of methods use pictorial structure model \cite{pictorial} for pose estimation, RoDG \cite{RoDG} leverages a pre-defined dependency graph representing relationships between adjacent body joints for pose estimation, the positions of these adjacent points are sequentially estimated along the dependency paths (skeleton path) from the root node. In deep learning era, the point-based methods \cite{fcos, centernet, centernet1, fsaf, cornernet, pointset} represent the instances by the grid points and have been applied in many tasks. They have drawn much attention as they are always simpler and more efficient than anchor-based representation \cite{fastrcnn, fasterrcnn, cascadercnn, maskrcnn}. CenterNet \cite{centernet} leverages bounding box center to encode the object information and regresses the other object properties such as size to predict bounding box in parallel. SPM \cite{spm} represents the person via root joint and further presents a fixed hierarchical body representation to estimate human pose. Point-Set Anchors \cite{pointset} propose to leverage a set of pre-defined points as pose anchor to provide more informative features for regression. In contrast to previous methods that use center or pre-defined pose anchor to model human instance, we propose to represent human instance via an adaptive point set including center and seven human-part related points as shown in Fig. \ref{fig:image3} (a). The novel representation is able to capture the diverse pose information and effectively model the connections between human instance and keypoints.


\section{Methodology}

First, we elaborate on the proposed body representation in subsection \ref{subsection3.1}. Then, subsection \ref{subsection3.2} gives a minute description of network architecture including Part Perception Module, Enhanced Center-aware Branch as well as Two-hop Regression Branch. Finally, we report the training and inference details in subsection \ref{subsection3.3}.

\begin{figure*}
\begin{center}
\includegraphics[width=2\columnwidth]{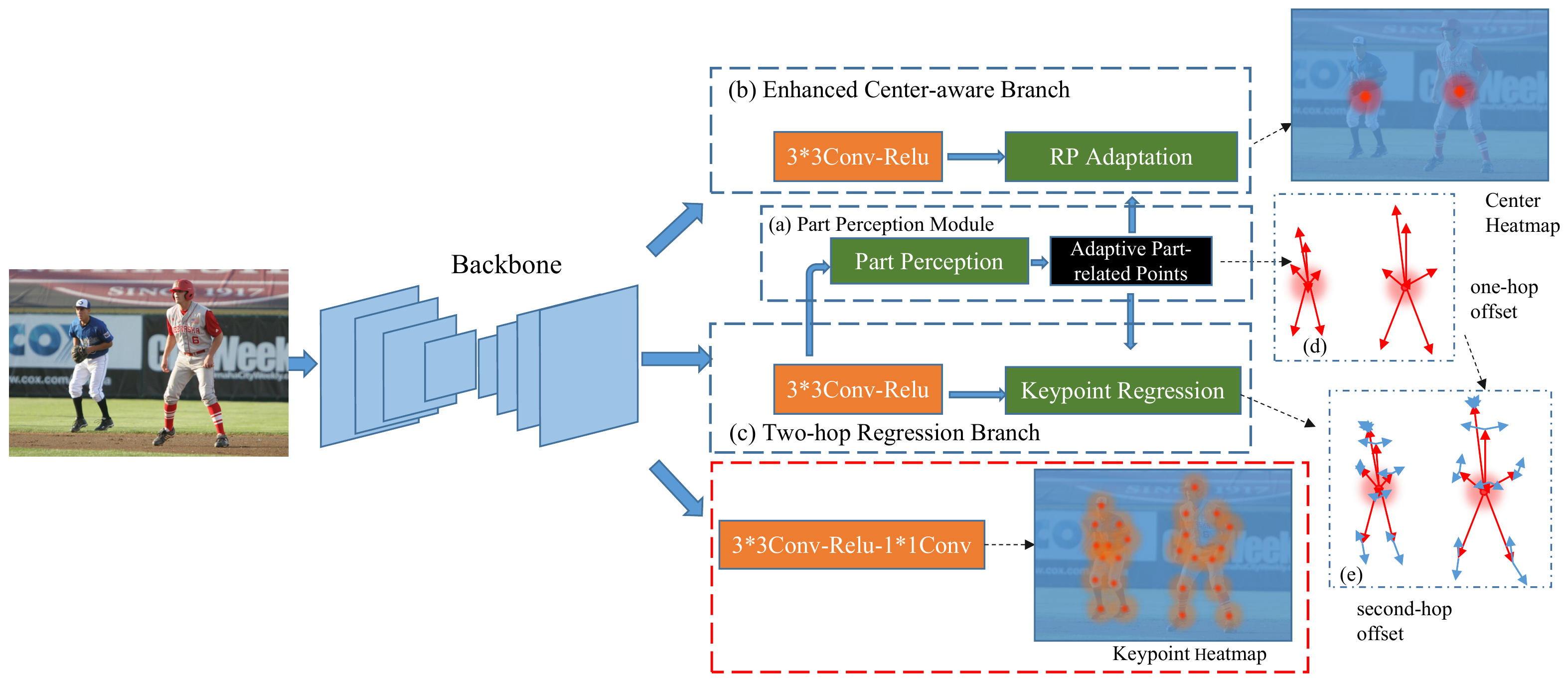}
\end{center}
\caption{ Overview of AdaptivePose.~(a) The structure of Part Perception Module.~(b) The structure of Enhanced Center-aware Branch, RP Adaptation refers to receptive field adaptation.~(c) The diagram of Two-hop Regression Branch.~(d) The red arrows are one-hop offsets that dynamically locate the adaptive human-part related points.~(e) The blue arrows indicate second-hop offsets for localizing the keypoints.}
\label{fig:network}
\end{figure*}

\subsection{Body Representation}\label{subsection3.1}
We present an adaptive point set representation that uses the center point together with several human-part related points to represent the human instance. The proposed representation introduces the adaptive human-part related points, whose features are used to encode the per-part information thus can sufficiently capture the structural pose information. Meanwhile, they serve as the intermediate nodes to effectively model the relationship between the human instance and keypoints. In contrast to the fixed hierarchical representation in SPM \cite{spm}, The adaptive part related points are predicted by center feature dynamically and not pre-defined locations, thus avoid the accumulated error propagated along the fixed hierarchical path. Furthermore, instead of only using the root feature to encode the whole pose information, the features of adaptive points are also leveraged to encode keypoint information of different parts more sufficiently in our method.



Our body representation is built upon the pixel-wise keypoint regression framework, which estimates the candidate pose at each pixel. For a human instance, we manually divide the human body into seven parts (i.e., head, shoulder, left arm, right arm, hip, left leg and right leg) according to the inherent structure of human body, as shown in Fig.~\ref{fig:image3} (b). Each divided human part is a rigid structure, we represent it via an adaptive human-part related point, which is dynamically regressed from the human center. The process can be formulated as:

\begin{equation} \label{one-hop}
  \mathbf{C}_{inst} \rightarrow \left \{\mathbf{P}_{head}, \mathbf{P}_{sho}, \mathbf{P}_{la}, \mathbf{P}_{ra}, \mathbf{P}_{hip}, \mathbf{P}_{ll}, \mathbf{P}_{rl}\right\},
\end{equation}
where $\mathbf{C}_{inst}$ refers to the instance center, others indicate seven adaptive human-part related points corresponding to head, shoulder, left arm, right arm, hip, left leg and right leg. Human pose is finely-grained represented by a point set $\left \{\mathbf{C}_{inst}, \mathbf{P}_{head}, \mathbf{P}_{sho}, \mathbf{P}_{la}, \mathbf{P}_{ra}, \mathbf{P}_{hip}, \mathbf{P}_{ll}, \mathbf{P}_{rl}\right\}$. By introducing the adaptive human-part related points, the semantic and position information of different keypoints can be encoded by the specific human-part related point's feature, instead of only using the limited center feature to encode all keypoints' information. For convenience, $\mathbf{P}_{part}$ is used to indicate the seven human-part related points $\mathbf{P}_{head}, \mathbf{P}_{sho}, \mathbf{P}_{la}, \mathbf{P}_{ra}, \mathbf{P}_{hip}, \mathbf{P}_{ll}, \mathbf{P}_{rl}$. Then the feature on each human-part related point is responsible for regressing the keypoints belong to corresponding parts as follows:

\begin{equation} \label{two-hop}
\mathbf{P}_{part} \rightarrow \mathbf{Joint}.
\end{equation}
The novel representation starts from the human center to adaptive human-part related points, then to body keypoints to build up the connection between the instance position and corresponding keypoint position in a single-forward pass without any non-differentiable process.

Based on the proposed representation, we deliver a single-stage differentiable solution to estimate multi-person pose. Concretely, Part Perception Module is proposed to predict seven human-part related points. By using the adaptive human-part related points, Enhanced Center-aware Branch is introduced to perceive the center of human with various deformation and scales. In parallel, Two-hop Regression Branch is presented to regress keypoints via the adaptive part-related points.


\subsection{Single-Stage Network}\label{subsection3.2}


{\noindent\bf Overall Architecture.} As shown in Fig. \ref{fig:network}, given an input image, we first extract the semantic feature via the backbone, following three well-designed components to predict specific information. We leverage Part Perception Module to regress seven adaptive human-part related points from the assumed center for each human instance. Then we conduct the receptive field adaptation in Enhanced Center-aware Branch by aggregating the features of the adaptive points to predict center heatmap. In addition, Two-hop Regression Branch adopts the adaptive human-part related points as one-hop nodes to indirectly regress the offsets from center to each keypoint. AdaptivePose follows pixel-wise keypoint regression paradigm, which estimates the candidate pose at each pixel (called center pixel), by predicting an 2K-dimensional offset vector from the center pixel to the K keypoints. We only take a pixel position as example to describe the single-stage network.

\begin{figure}
\begin{center}
\includegraphics[width=1.0\columnwidth]{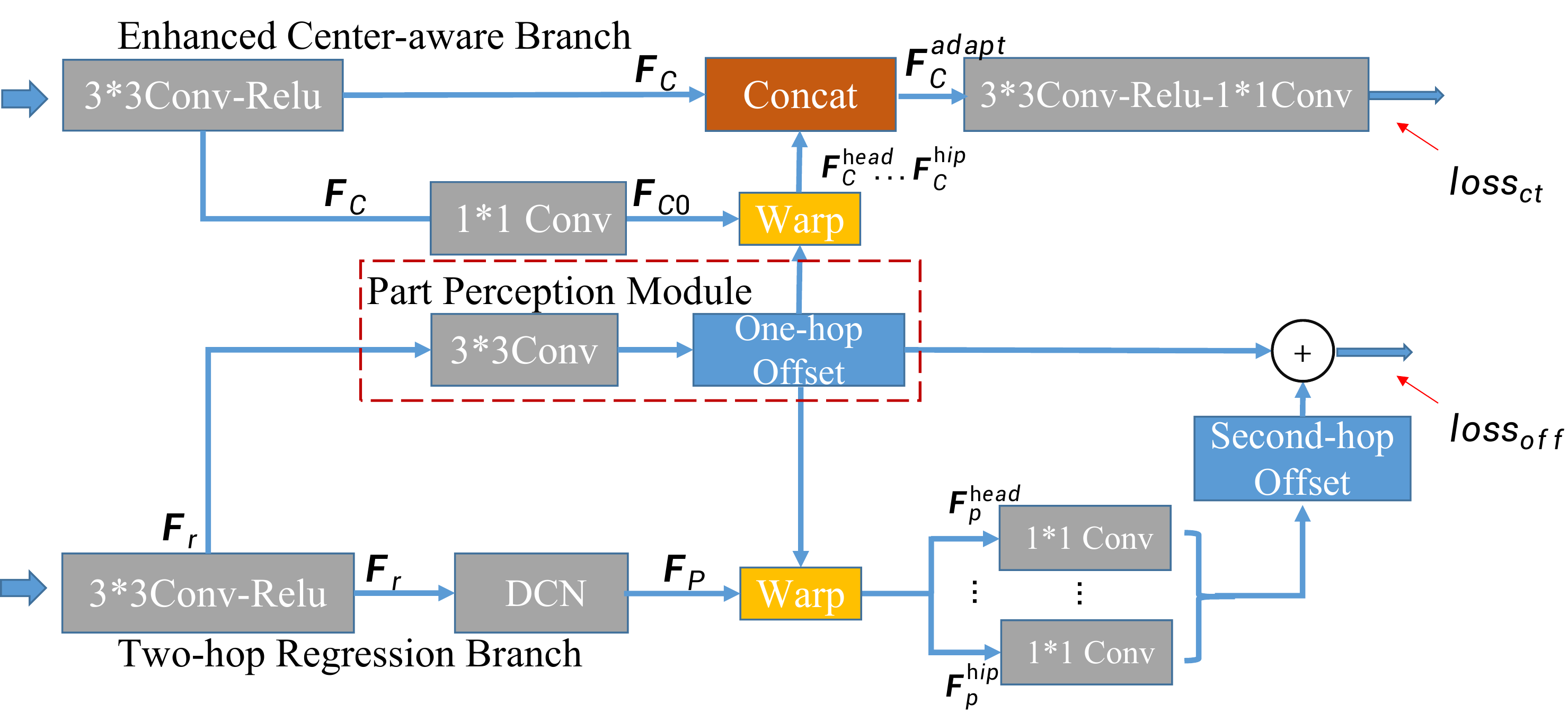}
\end{center}
\caption{The detailed structure across the Part Perception Module, Enhanced Center-aware Branch and Two-hop Regression Branch. Concat indicates the feature concatenation along the channel dimension. The red arrows is used to indicate where a loss is applied.}
\label{fig:detail}
\end{figure}

{\noindent\bf Part Perception Module.} With the proposed body representation, we artificially divide each human instance into seven local parts (i.e. head, shoulder, left arm, right arm, hip, left leg, right leg) according to the inherent structure of human body. Part Perception Module is proposed to perceive the human parts by predicting seven adaptive human-part related points. For each part, we automatically regress an adaptive point from center pixel $c$ without explicit supervision. Each adaptive part related point is considered as encoding the informative features for the keypoints belonging to this part. As shown in Fig. \ref{fig:detail}, we feed the regression branch specific feature $\mathbf{F}_{r}$ into the 3$\times$3 convolutional layer to regress 14-channel x-y offsets $\bar{\mathbf{off}_{1}}$ from the center $c$ to seven adaptive human-part related points on each pixel. These adaptive points act as intermediate nodes, which are used for subsequent center positioning and keypoint regression.

{\noindent \bf Enhanced Center-aware Branch.} In previous works \cite{centernet,spm,directpose}, the center of human instances with various scales and deformation are predicted via the features with fixed receptive field for each position. However, the pixel position which predicts the center of larger human body ought to have lager receptive field compared with the position for predicting the center of smaller human body. Thus we propose a novel Enhanced Center-aware Branch which consists of a receptive field adaptation process to extract and aggregate the features of seven adaptive human-part related points for precise center localization.


%


As shown in Fig.~\ref{fig:detail}, We use the structure of~3$\times$3 conv-relu to generate the branch-specific features. In Enhanced Center-aware Branch, $\mathbf{F}_{c}$ is branch-specific feature with the fixed receptive field for each pixel position. We firstly use the 1*1 convolution to compress 256-channel feature~$\mathbf{F}_{c}$ and obtain the 64-channel feature~$\mathbf{F}_{c0}$. Then we extract the feature vectors of the adaptive points via bilinear interpolation (named 'Warp' in Fig.~\ref{fig:detail}) on $F_{c0}$. Taking the head part as example, the bilinear interpolation can be formulated as 
$\mathbf{F}_{c}^{head} = \mathbf{F}_{c0}(c+ \bar{\mathbf{off}_{1}}^{head})$, where c indicates the center pixel and $\bar{\mathbf{off}_{1}}^{head}$ is offset from center to adaptive points of head part.
The extracted features \{$\mathbf{F}_{c}^{head}$,~$\mathbf{F}_{c}^{sho}$,~$\mathbf{F}_{c}^{la}$,~$\mathbf{F}_{c}^{ra}$,~$\mathbf{F}_{c}^{hip}$,~$\mathbf{F}_{c}^{ll}$,~$\mathbf{F}_{c}^{rl}$\} corresponds to seven divided human parts (i.e., head, shoulder, left arm, right arm, hip, left leg, right leg). We concatenate them with $\mathbf{F}_{c}$ along the channel dimension to generate the feature $\mathbf{F}_{c}^{adapt}$. Since the predicted adaptive points located on the seven divided parts are relatively evenly distributed on the human body region, thus the process above can be regarded as the receptive field adaptation according to the human scale as well as capture the various pose information sufficiently. Finally we use $\mathbf{F}_{c}^{adapt}$ with adaptive receptive field to predict the 1-channel probability map for the center localization.

We use the normalized Gaussian kernel $\mathbf{G}_{(x,y)}= \exp{(-\frac{(x-{C}_{x})^{2}+(y-{C}_{y})^{2}}{2*{\delta}^{2}})}$ with mean $({C}_{x},{C}_{y})$ and adaptive variance $ \delta $ calculated by human scale to generate the ground-truth center heatmap, following \cite{centernet, cornernet}. Concretely, we calculate the Gaussian kernel radius by the size of an object by ensuring that a pair of points within the radius would generate a bounding box with at least IoU 0.7 with the ground-truth annotation. The adaptive variance is 1/3 of the radius. For the loss function of the Enhanced Center-aware Branch, we employ the pixel-wise focal loss in a penalty-reduced manner as follows:

\begin{equation} \label{formula:1}
{loss}_{ct}= \frac{1}{N} \sum_{n=1}^{N} 
\left\{
\begin{aligned}
(1-\bar{P}_{c})^{\alpha}\;\ln{(\bar{P}_{c})} & & if {P}_{c}=1 \\
(1-{P}_{c})^{\beta} \; \bar{P}_{c}^{\alpha} \; \ln(1-\bar{P}_{c}) & & elif {P}_{c} \not= 1  ,\\
\end{aligned}
\right.
\end{equation}
where N refers to the number of positive sample, $\bar{P}_{c}$ and ${P}_{c}$ indicate the predicted per-pixel confidence and corresponding ground truth. $\alpha$ and $\beta$ are hyper-parameters and set to 2 and 4, following \cite{centernet, cornernet}. In above loss, only center pixels with peak 1.0 are positive samples and all others are negative samples.

{\noindent\bf Two-hop Regression Branch.}  We leverage a two-hop regression method to predict the displacements instead of directly regressing the center-to-joint offsets. In this manner, the adaptive human-part related points predicted by Part Perception Module act as one-hop nodes to build up the connection between human instance and keypoints more effectively.



We firstly leverage the structure of 3$\times$3 conv-relu to generate branch-specific feature, named $\mathbf{F}_{r}$ in Two-hop Regression Branch. Then we feed 256-channel $\mathbf{F}_{r}$ into the deformable convolutional layer \cite{dcn,dcnv2} to generate 64-channel feature $\mathbf{F}_{p}$. Then we extract the features at the adaptive part related points via the bilinear interpolation operation (called 'Warp' in Fig.~\ref{fig:detail}) on $\mathbf{F}_{p}$ for corresponding keypoint regression. We denote the extracted features as \{$\mathbf{F}_{p}^{head}$,~$\mathbf{F}_{p}^{sho}$,  ~$\mathbf{F}_{p}^{la}$,~$\mathbf{F}_{p}^{ra}$,~$\mathbf{F}_{p}^{hip}$,~$\mathbf{F}_{p}^{ll}$,~$\mathbf{F}_{p}^{rl}$\} corresponding to seven divided parts (i.e., head, shoulder, left arm, right arm, hip, left leg and right leg). For the extracted feature $\mathbf{F}_{p}^{head}$, the above process can be formulated as $\mathbf{F}_{p}^{head} = \mathbf{F}_{p}(c+ \bar{\mathbf{off}_{1}}^{head}).$ These features encode the keypoint information of different human parts respectively, instead of only using center features to encode the information of all keypoints. Afterward, the extracted features are responsible for locating the keypoints belonging to the corresponding part by regressing the offsets from adaptive part related
point to specific keypoints $\bar{\mathbf{off}_{2}}$ via different 1$\times$1 convolutional layers. For example, $\mathbf{F}_{p}^{head}$ is used to localize the eyes, ears and nose, $\mathbf{F}_{p}^{la}$ is used to localize the left wrist and left elbow, $\mathbf{F}_{p}^{ll}$ is used to localize the left knee and left ankle. 

Two-hop Regression Branch outputs a 34-channel tensor corresponding to x-y offsets $\bar{\mathbf{off}}$ from the center to 17 keypoints, which is predicted by the two-hop manner as follows:   

\begin{equation}
   \bar{\mathbf{off}} = \bar{\mathbf{off}_{1}}+\bar{\mathbf{off}_{2}},
\end{equation}
where $\bar{\mathbf{off}_{1}}$ and $\bar{\mathbf{off}_{2}}$ respectively indicate the offset from the center to adaptive human-part related point~(One-hop offset mentioned in Fig.~\ref{fig:detail}) and the offset from human-part related point to specific keypoints~(second-hop offset mentioned in Fig.~\ref{fig:detail}). The predicted offsets $\bar{\mathbf{off}}$ are supervised by vanilla L1 loss and the supervision only acts at positive keypoint locations, the other background locations are ignored. Furthermore, we add an additional loss term to learn the rigid bone connection between adjacent keypoints, termed Skeleton-Aware Regression Loss. In particular, as shown in Fig. \ref{fig:image3}(c), we denote a bone connection set as $\mathcal{B} = \{\textbf{B}^{i}\}_{i=1}^{I}$, where $I$ is the number of bone connection in pre-defined set $\mathcal{B}$. Each bone is formulated as $\textbf{B} = \textbf{P}_{adjacent(joint)} - \textbf{P}_{joint}$, in which $\textbf{P}$ is the joint position and the function $adjacent(*)$ return the adjacent joints for input joint. The total regression loss is formulated as follows:
\begin{equation}
   {loss}_{off}= \frac{1}{2K*N} \sum_{n=1}^{K} \left|  \bar{\mathbf{off}^{n}}- \mathbf{off}^{n}_{gt} \right| + \frac{1}{2I*N} \sum_{i=1}^{I} \left|  \bar{\textbf{B}^{i}}- \textbf{B}^{i}_{gt} \right|,
\label{formula:2}
\end{equation}
where $\mathbf{off}^{n}_{gt}$ and $\mathbf{B}^{i}_{gt}$ are the ground truth center-to-keypoint offset and bone connection. $N$ indicates the number of human instances. $K$ is the number of valid keypoint locations. We find that employing the supervision on the bone connections of inner parts and cross parts can bring 0.3 AP improvements on CrowdPose \cite{crowdpose}.



\subsection{Training and Inference Details}\label{subsection3.3}



During training, we employ an auxiliary training objective to learn keypoint heatmap representation, which enables the feature to maintain more human structural geometric information. In particular, we add a parallel branch to output a 17-channel heatmap corresponding to 17 keypoints and apply a Gaussian kernel with adaptive variance to generate ground truth keypoint heatmap. We denote this training objective as ${loss}_{hm}$, which is similar to Equation \ref{formula:1}. The only difference is that N refers to the number of positive keypoints. The auxiliary branch is only used for training process and removed in inference process.  

Our total training loss for multi-task training procedure is formulated as:
\begin{equation}
   {loss}_{total}= {loss}_{ct} + {loss}_{off} + {loss}_{hm}.
\end{equation}

During inference, Enhanced Center-aware Branch outputs the center heatmap that indicates whether the pixel position is center or not. Two-hop Regression Branch outputs the offsets from the center to each keypoint. We firstly pick the human center by using 5$\times$5 max-pooling kernel on the center heatmap to maintain 20 candidates, and then retrieve the corresponding offsets $(\delta_{x}^{i},\delta_{y}^{i})$ to form human pose without any extra tricks. Specifically, we denote the predicted center as $({C}_{x},{C}_{y})$. The above decode process is formulated as follows:

\begin{equation}
   ({K}_{x}^{i},{K}_{y}^{i})= ({C}_{x},{C}_{y}) + (\delta_{x}^{i},\delta_{y}^{i}) ,
\end{equation}
where $({K}_{x}^{i},{K}_{y}^{i})$ is the coordinate of the i-th keypoint. In contrast to DEKR\cite{dekr} that further uses the average of the extracted heat values at each regressed keypoints to modulate the center heat-values, we only leverage the center heat-values as the final pose score for fast inference.


\section{Experiments and Analysis}

In this section, we first briefly introduce the 2D pose estimation datasets, evaluation metric, data augmentation and implementation details in subsection~\ref{subsection4.1}. Next, we conduct comprehensive ablation studies to reveal the effectiveness of each component in subsection~\ref{subsection4.2}. Then, we compare our proposed method with the previous methods on MS COCO \cite{coco} in subsection~\ref{subsection4.3} and CrowdPose \cite{crowdpose} in subsection~\ref{subsection4.4}. Finally, we extent our network to 3D multi-person pose estimation and verify the generalizability on 3D MuCo-3DHP \cite{muco} and MuPoTS-3D \cite{muco} dataset.

\subsection{Experimental Setup}\label{subsection4.1}

\noindent{\bf Dataset.}  We evaluate our proposed AdaptivePose framework on two 2D
multi-person pose estimation benchmarks including MS COCO \cite{coco} and CrowdPose \cite{crowdpose}. MS COCO dataset~\cite{coco} is a large-scale pose estimation benchmark consists of over 200,000 images for more than 250k human instances annotated with 17 keypoints. It is divided into train, validation, test set respectively. We train our model on COCO train2017 dataset. The comprehensive experimental results are reported on the COCO mini-val set with 5000 images and test-dev2017 set with 20K images. The CrowdPose \cite{crowdpose} dataset consists of 20,000 images for 80,000 labelled persons. The training, validation and test set are partitioned in proportional to 5:1:4. It contains more challenging images which used to verify the robustness for crowded scenes.  We follow previous works \cite{higherhrnet, dekr,centergroup,luo} that train our models on the train and validation sets and report the results on the test set.

\noindent{\bf Evaluation Metric.} We leverage average precision and average recall based on different Object Keypoint Similarity~(OKS)\cite{coco} thresholds to evaluate our keypoint detection performance on both MS COCO and CrowdPose dataset. OKS is formulated as follows:

\begin{equation}
   {OKS}= \frac{\sum_{i}exp(\frac{-{d}_{i}^{2}}{2{s}^{2}{k}_{i}^{2}}) \delta(\upsilon_{i}>0)}    {\sum_{i}\delta(\upsilon_{i}>0)},
\end{equation}
where $d_{i}$ is the Euclidean distance between the predicted keypoint and the corresponding ground-truth, $\upsilon_{i}$ represents the visibility tag of keypoint,
$\delta$ in a function when $\upsilon_{i}>0$ is 1, otherwise is 0, $s$ refers to the instance scale, and $k_{i}$ is a constant to control falloff for each specific keypoint. In addition, for COCO dataset, we report AP$_{M}$ and AP$_{L}$, which corresponds to AP over medium and large-sized instances respectively. For CrowdPose, we report AP$_E$, AP$_M$, AP$_H$, which indicate AP scores over easy, medium and hard instances according to dataset annotations.

\noindent{\bf Data Augmentation.} During training, we use random flip, random rotation, random scaling and color jitter to augment training samples. The flip probability is set to 0.5, the rotation range is (-30, 30) and the scale range is (0.6, 1.3). Each input image is cropped according to the random center and random scale and then resized to 512 / 640 / 800 pixels for different backbones.

\noindent{\bf Implementation Details.} We train our proposed model via Adam \cite{adam} optimizer with initial learning rate of 2.5e-4 on the workstation with eight Tesla V100 GPUs. The learning rate is dropped to 2.5e-5 and 2.5e-6 at the 230th and 260th epochs, respectively. The total training procedure is terminated at 280th epoch (2$\times$ training scheme). All codes are implemented with Pytorch. DLA-34~(19.7M) \cite{dla} and HRNet \cite{hrnet} are adopted to achieve the trade-offs between the accuracy and efficiency. The batch size is set to 128 for DLA-34 and HRNet-W32 and 64 for HRNet-W48 due to the limited GPU memory. During inference, we keep the aspect ratio of the raw image and resize the short side of the images to 512 / 640 / 800 pixels. The output size is 1/4 of the input resolution. We further use flip and multi-scale image pyramids to boost the performance. It is worth highlighting that the flip is only applied to the center heatmap predicted by Enhanced Center-aware Branch. All training and inference setups are shared between MS COCO \cite{coco} and CrowdPose \cite{crowdpose} dataset. 

\begin{table}
\begin{center}
{\caption{Ablation study for exploring the design of Part Perception Module. }\label{tab:PPM}}
\resizebox{0.95\columnwidth}{!}{
\begin{tabular}{l|cccccc}
\hline
Structure  &  $AP$ & $AP_{50}$ & $AP_{75}$ & $AP_{M}$ & $AP_{L}$   \\

\hline

1$\times$1 Conv & 64.5 & 85.9 & 70.6 & 58.0 & 73.9 \\
Group 3$\times$3 Conv &64.5& 85.6 & 70.6 & 57.9& 74.0 \\
Trident 3$\times$3 Conv &64.9& 86.1 & 70.9 & 58.3& 74.3 \\
vanilla 3$\times$3 Conv &64.6& 85.5 & 70.4 & 58.1& 74.2 \\

\hline

\end{tabular}
}
\end{center}
\vspace{-2mm}
\end{table}

\subsection{Ablation Experiments}\label{subsection4.2}
In this subsection, we conduct comprehensive ablation experiments to analyze each component respectively as well as our whole regression model. All ablation studies adopt DLA-34 as backbone and use the 1$\times$ training schedule (140 epochs) via single-scale testing without horizontal flip on the COCO mini-val set.

\begin{figure*}
\begin{center}
\includegraphics[width=1.9\columnwidth]{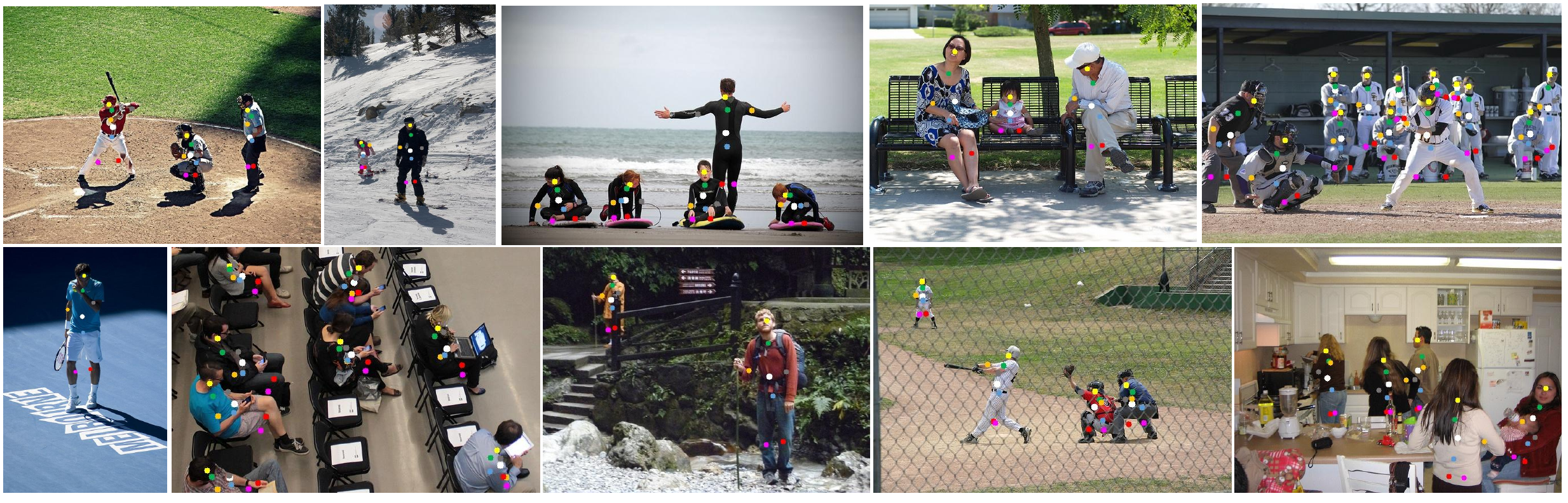}
\end{center}
\caption{(a) Visualization of the predicted adaptive point set including the center and seven human-part related points for each human instance on COCO dataset. Best viewed after zooming in.}
\label{fig:adaptivepoint}
\vspace{-3mm}
\end{figure*}

\noindent{\bf Analysis of Part Perception Module.}
Based on the adaptive point set representation, Part Perception Module is proposed to regress seven adaptive human-part related points, which are used for subsequent prediction of Enhanced Center-aware Branch and Two-hop Regression Branch. Fig. \ref{fig:adaptivepoint} shows the predicted human center and seven adaptive human-part related points on the human instances with various scales and pose.

As reported in Table~\ref{tab:PPM}, we leverage various designs to study the structure of Part Perception Module including (1) 1$\times$1 convolutional layer; (2) 3$\times$3 convolutional layer with group 7, in which each group is responsible for a human part; (3) trident 3$\times$3 convolutional layer with dilation rate 1, 2, 3 respectively; (4) vanilla 3$\times$3 convolutional layer. The trident convolution achieves the slightly better result in 1x training scheme. However, with the 2x training scheme, the vanilla 3$\times$3 convolution obtain the equal performance. We select the vanilla 3$\times$3 convolution for the follow-up experiments.


\noindent{\bf Analysis of Enhanced Center-aware Branch.} In Enhanced Center-aware Branch, we conduct the receptive field adaptation operation by aggregating the feature vectors of seven adaptive human-part related points to more precisely positioning the human center.

We conduct the controlled experiments to explore the effect of receptive field adaptation (RFA) process in Enhanced Center-aware Branch. Compared with using the feature with fixed receptive field to positioning the human center, receptive field adaptation process obtains 1.4\% AP improvements in ($Expt.$~3 versus $Expt.$~4) of Table~\ref{tab:ablation}. We consider that receptive field adaptation is capable of enhancing center feature representation and dynamically adjusting its receptive field accordingly.

\begin{table}
\begin{center}
{\caption{Ablation studies. $PPM$ denotes Part Perception Module, $RFA$ is receptive field adaptation conducted in Enhanced Center-aware Branch, $TR$ indicates two-hop regression strategy in Two-hop Regression Branch, $AL$ refers to employ an auxiliary loss ${loss}_{hm}$ to learn keypoint heatmap representation.}\label{tab:ablation}}
\resizebox{1\columnwidth}{!}{
\begin{tabular}{l|cccc|ccccc}
\hline
$ Expt.$ & $PPM$ & $RFA$ & $TR$ & $AL$ & $AP$ & $AP_{50}$ & $AP_{75}$ & $AP_{M}$ & $AP_{L}$  \\
\hline
$1$&- & -&- & -&  57.5 & 81.7 & 62.6 & 48.7 & 70.2 \\
$2$&$\surd$ & $\surd$ & -& -& 58.5 & 82.4 & 63.8 & 49.0 & 71.7 \\
$3$&$\surd$ & - &$\surd$ & -& 61.6 & 84.5 & 65.9 & 53.0 & 72.1  \\                        
$4$ & $\surd$ & $\surd$ &$\surd$ &- & 63.0 & 85.5&68.6&56.0&72.8 \\

$5$&$\surd$ & $\surd$ &$\surd$ & $\surd$ & {\bf64.6} & {\bf85.5} & {\bf70.4} & {\bf58.1} & {\bf74.2}  \\
\hline

\end{tabular}}

\end{center}
\vspace{-1mm}
\end{table}

\noindent{\bf Analysis of Two-hop Regression Branch.} In Two-hop Regression Branch, we adopt the adaptive human-part related points as intermediate nodes to localize the keypoints along the center-to-adaptive points-to-keypoints path.

In order to investigate the effect of adaptive two-hop regression method, As reported in ($Expt.$~2 versus $Expt.$~4) of Table~\ref{tab:ablation}. It brings to 4.5\% AP improvements compared with directly regressing the displacements from center to each joints. The results prove the feature embedding of adaptive point is capable of sufficiently encoding the content and position information of corresponding keypoints than limited center feature embedding. Thus these adaptive points serve as the intermediate nodes can factorize center-to-joint offsets effectively to lift the regression performance. 

Furthermore, we analyze the localization error of the direct center-to-joint regression, hierarchical regression in SPM \cite{spm} and our adaptive two-hop regression (with auxiliary keypoint heatmap loss applied for three above regression schemes) via the coco-analyze tool \cite{cocoanalyse}. The localization error consists of four error types: (1) $Jitter$ is small localization error; (2) $Miss$ refers to large localization error; (3) $Inversion$ denotes confusion between keypoints within an human instance; (4) $Swap$ indicates the confusion between keypoints across different human bodies. The results are shown in Table \ref{tab:TR}, compared with direct center-to-joint regression and hierarchical regression, our adaptive two-hop regression reduces $Jitter$ error by 4.5 and 1.4 respectively as well as reduces $Miss$ error by 1.9 and 0.5, which proves that our regression method can improve the localization quality of the other regression methods evidently.

\begin{figure*}
\begin{center}
\includegraphics[width=1.9\columnwidth]{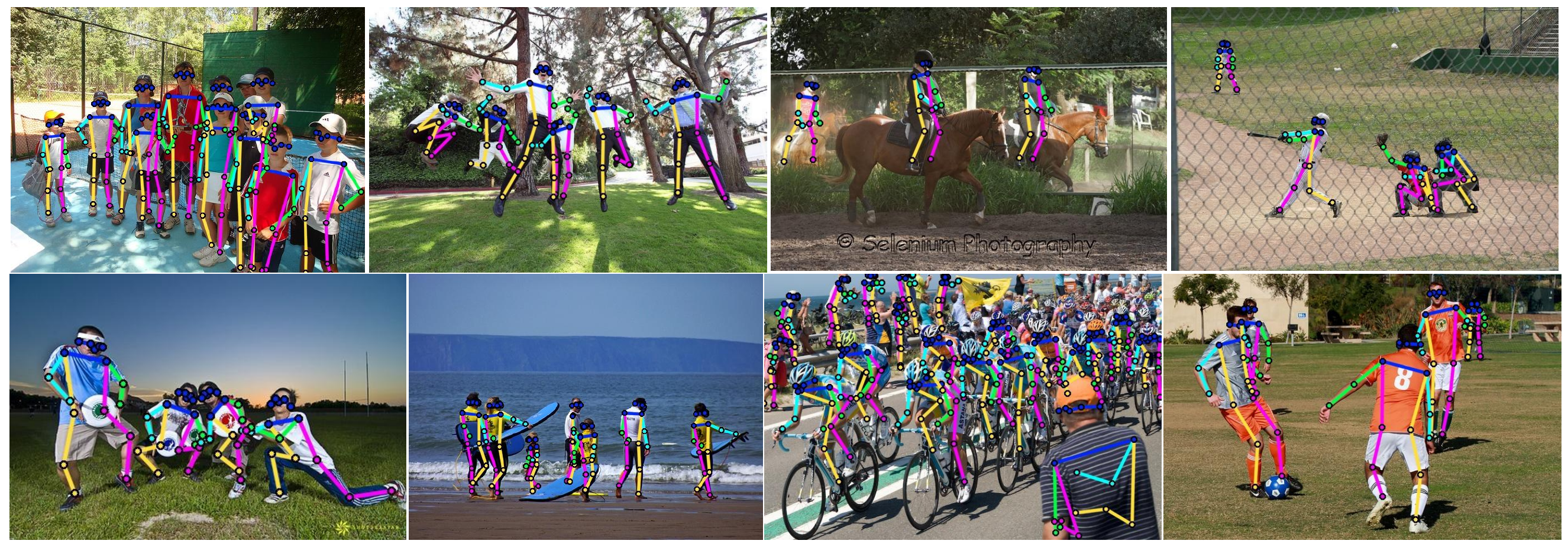}
\end{center}
\caption{Examples of predicted skeleton on COCO mini-val dataset. The challenges include various human scales, pose deformation as well as occluded scenarios. Best viewed after zooming in.}
\label{fig:coco_vis}
\vspace{-1mm}
\end{figure*}

\noindent{\bf Analysis of auxiliary loss.} We add a parallel branch to learn keypoint heatmap representation, which is only used for auxiliary loss computation in training stage.

($Expt.$~4 versus $Expt.$~5) of Table~\ref{tab:ablation}
study the effect of auxiliary loss, we achieve 1.6\% AP improvements by employing auxiliary heatmap loss to help coordinate regression. It experimentally proves that learning the keypoint heatmap is able to retain more structural geometric information to improve regression performance.

\begin{table}
\begin{center}
{\caption{Comparisons of direct center-to-joint regression (denoted as DR) in CenetrNet, hierarchical regression (denoted as HR) in SPM and our Two-hop Regression (denoted as TR) in terms of the precision and four types of localization errors.}\label{tab:TR}}
\resizebox{0.9\columnwidth}{!}{
\begin{tabular}{l|c|cccc}
\hline
 Methods &  AP $\uparrow$ & $Jitter\downarrow$ & $Miss\downarrow$ & $Inversion \downarrow$ & $Swap \downarrow $  \\

\hline

DR & 58.8 & 17.3 & 9.5 & 3.9 & 1.2 \\

HR & 62.2 & 14.2 & 8.1 & 3.8& {\bf 1.2} \\

TR & {\bf 63.8} & {\bf 12.8} & {\bf 7.6} & {\bf 3.5}& 1.3 \\

\hline

\end{tabular}
}
\end{center}
\vspace{-2mm}
\end{table}

\noindent{\bf Analysis of Overall Architecture.} We study the inherent relationship between Enhanced Center-aware Branch and Two-hop Regression Branch, which are correlated by the adaptive human-part related points. As shown in $Expt.$~1 and 2 of Table~\ref{tab:ablation}, without two-hop regression, receptive field adaptation achieves 1.0\% AP improvements. As reported in $Expt.$~3 and 4 of Table~\ref{tab:ablation}, with two-hop regression, we further observe that receptive field adaptation achieves 1.4\% AP improvements. We consider that ${loss}_{off}$ enables the adaptive points to scatter over the divided human parts, thus the receptive field adaptation is capable of perceiving the human center more precisely. Meanwhile, as reported in $Expt.$~1 and 3 of Table~\ref{tab:ablation}, without receptive field adaptation, two-hop regression brings 4.1\% AP improvements. With receptive field adaptation, two-hop regression brings 4.5\% AP improvements as shown in $Expt.$~2 and 4 of Table~\ref{tab:ablation}. It experimentally proves that ${loss}_{ct}$ drives the adaptive points to locate on the semantically significant region, thus two-hop regression is able to better locate the keypoints.

\begin{table}
\begin{center}
{\caption{Ablation study for exploring the effect of heatmap refinement for our two-hop regressed result. }\label{tab:heat}}
\resizebox{0.8\columnwidth}{!}{
\begin{tabular}{lcccccc}
\hline
 Methods &  $AP$ & $AP_{50}$ & $AP_{75}$ & $AP_{M}$ & $AP_{L}$   \\

\hline

Ours-heat & 64.4 & 85.3 & 70.3 & 58.3 & 73.8 \\

Ours-reg &64.6& 85.5 & 70.4 & 58.1& 74.2 \\

\hline

\end{tabular}
}
\end{center}
\vspace{-2mm}
\end{table}

\noindent{\bf Analysis of Heatmap Refinement.} CenterNet~\cite{centernet} performs a post-processing step, which searches the closest peaks~(confidence \textgreater~0.1) on the keypoint heatmap to replace the initial regressed results. Since the position of confidence peaks on keypoint heatmap are integer, thus sub-pixel offsets are predicted to recover the discretization errors in parallel. In this manner, the regressed predictions are grouping clues for assigning the keypoints detected from heatmap to individuals. We named the above process as heatmap refinement.

\begin{table*}
\begin{center}
{\caption{Comparisons with the competitive bottom-up methods on the COCO mini-val set. Note that all results are reported for {\textbf single-scale testing without horizontal flip}. $^{\ddagger}$ indicates the conference version. + refers to larger input resolution. All inference time is calculated on a 2080Ti GPU with mini-batch 1.} \label{tab:val}}
\resizebox{1.8\columnwidth}{!}{

\begin{tabular}{l|c|c|c|c|cc|cc|c}
\hline
Methods& Params.(M) &Input Size & Output Size& $AP$ & $AP_{50}$ & $AP_{75}$ & $AP_{M}$ & $AP_{L}$ &Time(s) \\  

\hline
\hline
\multicolumn{10}{c}{Most Competitive Bottom-Up Methods}\\
\hline



DEKR-W32~\cite{dekr} &29.6  &512 & 128 & 63.4 & 86.2 & 69.8 & 55.8 & {\bf76.2}& 0.145  \\

HrHRNet-W32~\cite{higherhrnet} &28.6  &512 & 256 & 63.6 & 84.9 & 69.2 & 57.1 & 73.5 & 0.164  \\
SWAHR-W32~\cite{luo}  &28.6  &512 & 256 & 64.7 & 86.1 & 69.8 & 57.8 & 74.8 &0.235 \\

CenterGroup-W32~\cite{centergroup}  &30.3  &512 & 256  & 66.9 & - & - & - & -& - \\

\hline
\multicolumn{10}{c}{Single-Stage Methods}\\
\hline
AdaptivePose$^{\ddagger}$ (DLA-34)~\cite{adaptivepose} &21.0  &512 & 128 & 65.5 & 86.1 & 71.9 & 59.1 & 75.0 &{\bf0.034}   \\
\hline
Ours~(DLA-34) &21.0  &512 & 128 & 65.8 & 86.4 & 71.9 & 59.3 & 75.4 &{\bf0.034}   \\
Ours~(HRNet-W32) &29.6  &512 & 128 & {\bf68.0} & {\bf88.1} & {\bf74.2} & {\bf62.2} & 75.6 &0.056  \\

\hline
\hline
\multicolumn{10}{c}{Most Competitive Bottom-Up Methods+}\\
\hline
Li et al.~\cite{csvt1} (Hourglass-104) & -&640&160&66.0&-&-&-&-&-   \\
DEKR-W48~\cite{dekr} &65.7  &640 & 160 & 67.1 & 87.7 & 73.9 & 61.5 & 77.1&0.195  \\

HrHRNet-W48~\cite{higherhrnet} &63.8  &640 & 320 & 66.6 & 85.3 & 72.8 & 61.7 & 74.4&0.242  \\
SWAHR-W48~\cite{luo}  &63.8  &640 & 320 & 67.3 & 87.1 & 72.9 & 62.1 & 75.0 & 0.428 \\

CenterGroup-W48~\cite{centergroup}  &65.5  &640 & 320 &  69.1 & - & - & - & - &- \\

\hline
\multicolumn{10}{c}{Single-Stage Methods+}\\
\hline
AdaptivePose$^{\ddagger}$~(DLA-34)~\cite{adaptivepose}&21.0  &640 & 160 & 66.6 & 86.0 & 72.2 & 60.1 & 75.8  &{\bf0.045} \\
\hline
Ours~(DLA-34)&21.0  &640 & 160 & 67.0 & 86.3 & 72.6 & 60.5 & 76.1  &{\bf0.045} \\


Ours~(HRNet-W48) &64.8  &640 & 160 & 70.5& {\bf88.5} & 76.7 & 64.5& {\bf79.4} &0.082 \\
\bf{Ours~(HRNet-W48)} &64.8  &800 & 200 & {\bf70.8} &  88.3 & {\bf77.0} & {\bf65.7} & 78.7 & 0.115  \\
\hline

\end{tabular}
}
\end{center}
\vspace{-2mm}
\end{table*}


As reported in CenterNet~\cite{centernet}, heatmap refinement brings a large improvements of 6.2\% AP to the initial regression result~(from 51.7\% AP to 57.9\% AP). For validating the regression performance of our method, we further conduct the heatmap refinement to our regression result. For convenience, the two-hop regression result and the heatmap refinement result are denoted as Ours-reg and Ours-heat respectively. As shown in Table~\ref{tab:heat}, Ours-reg obtain slightly better performance than Ours-heat~(64.6\% AP versus 64.4\% AP), which proves that our regression method has the better positioning capacity.


\subsection{Results on MS COCO Dataset}\label{subsection4.3}
We report the comparisons with the previous state-of-the-art methods on COCO mini-val and test-dev set. All experimental results are obtained via 2x training schedule.

\noindent{\textbf{Mini-val Results.}} Table~\ref{tab:val} reports the comparisons with recent most competitive bottom-up methods to reveal the keypoint positioning capability in case of without any test-time augmentation (single-scale testing without flip). Adopting smaller DLA-34 as backbone and the same input resolution 512 pixels, our method achieves 65.8 AP, which outperforms competitive bottom-up HrHRNet-W32 \cite{higherhrnet}, SWAHR-W32 \cite{luo} as well as DEKR-W32 \cite{dekr} by 2.2 AP, 1.1 AP, 2.4 AP respectively with much faster inference speed. By using HRNet-W32, we outperform state-of-the-art CenterGroup-W32 \cite{centergroup} by 1.1 \% AP. Adopting DLA-34 and 640 pixels input resolution, our network achieves the equal performance with HrHRNet-W48, SWAHR-W48 as well as DEKR-W48 with only 1/3 parameters. Furthermore, we obtain 70.5 AP by using HRNet-W48 with 640 pixels input resolution, which achieve 3.4 AP gain over state-of-the-art regression-based method DEKR-W48. It is noteworthy that DEKR leverages the keypoint heat-value to modulate the center heat-value and further employs extra rescoring network in post-process, while we directly adopt the center heat-value as final pose score for fast inference. For the state-of-the-art bottom-up method CenterGroup-W48 (adopting HigherHRNet-W48 as backbone) which introduces transformer encoder to conduct grouping process, we surpass it by 1.4 AP via HRNet-W48 without extra deconvolution layers. The above results prove that our proposed body representation is more effective to model the relationship between human instance and kepoints than previous heuristic or learnable grouping methods in terms of accuracy and speed. Fig. \ref{fig:coco_vis} shows the predicted skeletons on COCO mini-val set.

\begin{table}
\begin{center}
{\caption{Comparisons with previous state-of-the-art methods on COCO test-dev set. ${*}$ indicates using extra test-time refinements. $\dag$ refers to multi-scale testing. DLA-34+ indicates DLA-34 with 640 pixels input resolution.}\label{tab:test}}
\resizebox{1\columnwidth}{!}{

\begin{tabular}{l|c|ccccc}
\hline
Methods & Params.(M) & $AP$ & $AP_{50}$ & $AP_{75}$ & $AP_{M}$ & $AP_{L}$  \\  
\hline

\hline
\multicolumn{7}{c}{Bottom-Up Methods}\\
\hline
CMU-Pose$^{*\dag}$~\cite{openpose} &-&61.8& 84.9& 67.5& 57.1 &68.2 \\

AE$^{*\dag}$~\cite{ae} &227.8 & 65.5 & 86.8 & 72.3 & 60.6 & 72.6  \\
CenterNet-DLA~\cite{centernet} &- &57.9&84.7&63.1&52.5&67.4\\
CenterNet-HG~\cite{centernet} &- & 63.0 & 86.8 & 69.6 & 58.9 & 70.4 \\
PersonLab~\cite{personlab} &68.7 & 66.5& 88.0& 72.6& 62.4& 72.3 \\
PifPaf~\cite{pifpaf} & - & 66.7 & - & -& 62.4& 72.9 \\
HrHRNet-W48$^{*\dag}$~\cite{higherhrnet} & 63.8& 70.5 & 89.3 & 77.2 & 66.6&75.8   \\
FCPose-R101 \cite{fcpose} & -&65.6&87.9&72.6&62.1& 72.3\\
SWAHR-W48$^{*}$ \cite{luo} &63.8&70.2&89.9&76.9&65.2&77.0\\
DEKR-W48$^{* \dag}$~\cite{dekr} & 65.7 & 71.0 & 89.2 & 78.0 & 67.1& 76.9 \\

\hline
\multicolumn{7}{c}{Single-stage Regression-based Methods}\\
\hline
SPM $^{* \dag}$~\cite{spm} & - & 66.9& 88.5& 72.9& 62.6& 73.1 \\
DirectPose $^{\dag}$~\cite{directpose} &- &64.8& 87.8& 71.1& 60.4& 71.5 \\
PointSetNet $^{* \dag}$~\cite{pointset} &- &68.7& 89.9& 76.3& 64.8& 75.3 \\

\hline

Ours~(DLA-34)$^{\dag}$ &21.0& 67.5 & 88.3 & 73.7 & 62.7 & 74.5 \\
Ours~(DLA-34+)$^{\dag}$ &21.0& 68.4 & 88.9 & 75.5 & 63.6 & 75.4  \\
\bf{Ours~(HRNet-W48)}$^{\dag}$ & 64.7 & {\bf71.4} & {\bf90.2} & {\bf78.5} & {\bf66.8} & {\bf78.2}   \\
\hline
\end{tabular}
}

\end{center}
\end{table}



\noindent{\textbf{Test-dev Results.}}~We further list comprehensive comparisons with the existing bottom-up and single-stage regression-based methods on COCO test-dev set. In details, as reported in Table~\ref{tab:test}, our method achieves state-of-the-art 71.4 AP, which outperforms the widely-used bottom-up methods CMU-pose~\cite{openpose} and AE~\cite{ae} by a large margin with faster inference speed. Finally, compared with previous single-stage regression-based methods, our method surpasses SPM~\cite{spm}~(refined by the well-trained single-person pose estimation model) by 4.5 AP without any refinement and also outperforms DirectPose~\cite{directpose} with a large margin by 5.1 AP.

\subsection{Results on CrowdPose Dataset}\label{subsection4.4}
We compare our AdaptivePose with the previous state-of-the-art methods on CrowdPose dataset which consists of more challenging crowd scenes. Following existing methods \cite{dekr, higherhrnet, luo}. We train our models on the train and val sets and evaluate the performance on the test set.

The comparisons with previous state-of-the-art methods are shown in Table \ref{tab:crowdpose}. Generally, the top-down paradigm always achieve the better performance than bottom-up and single-stage paradigm due to the persons are cropped to perform single-person pose estimation. Nevertheless, our single-stage methods achieve the better performance than most widely-used top-down methods on CrowdPose. We consider that the detected single person region always contains the bodies of other persons in crowd scenes, where persons are usually heavily overlapped. 

\begin{figure*}
\begin{center}
\includegraphics[width=1.9\columnwidth]{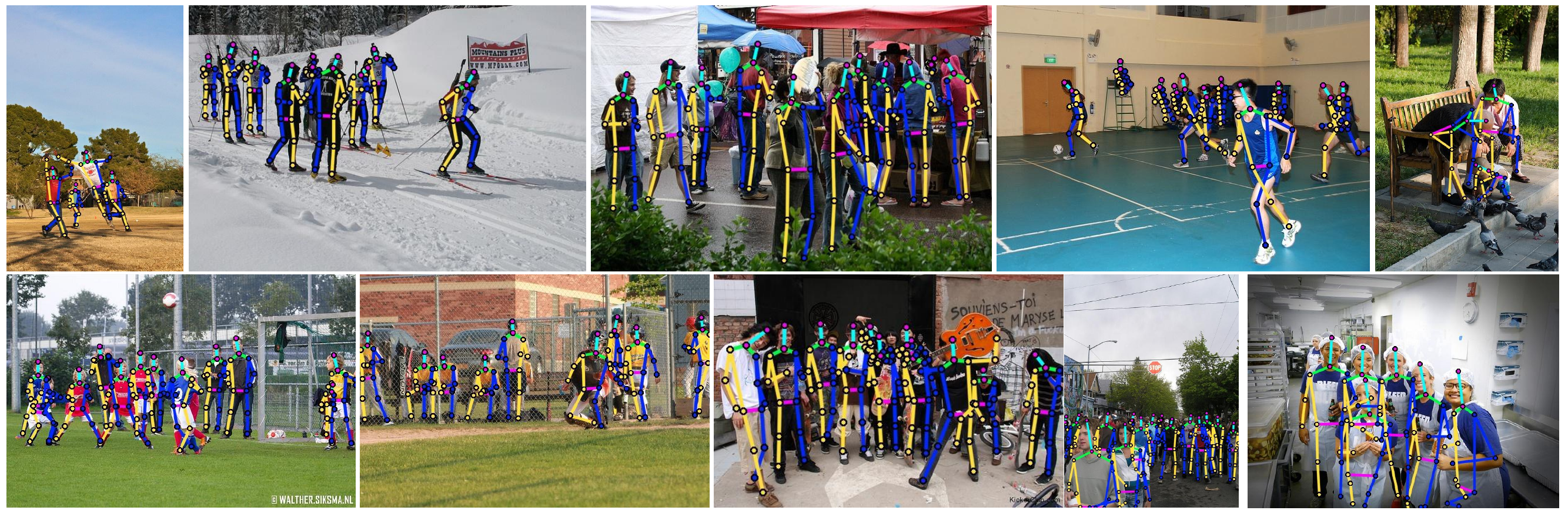}
\end{center}
\caption{Visualization of the estimated multi-person pose on CrowdPose dataset. The examples consist of many overlapped and occluded multi-person scenarios. Best viewed after zooming in.}
\label{fig:crowdpose_vis}
\end{figure*}

\begin{figure*}
\begin{center}
\includegraphics[width=1.9\columnwidth]{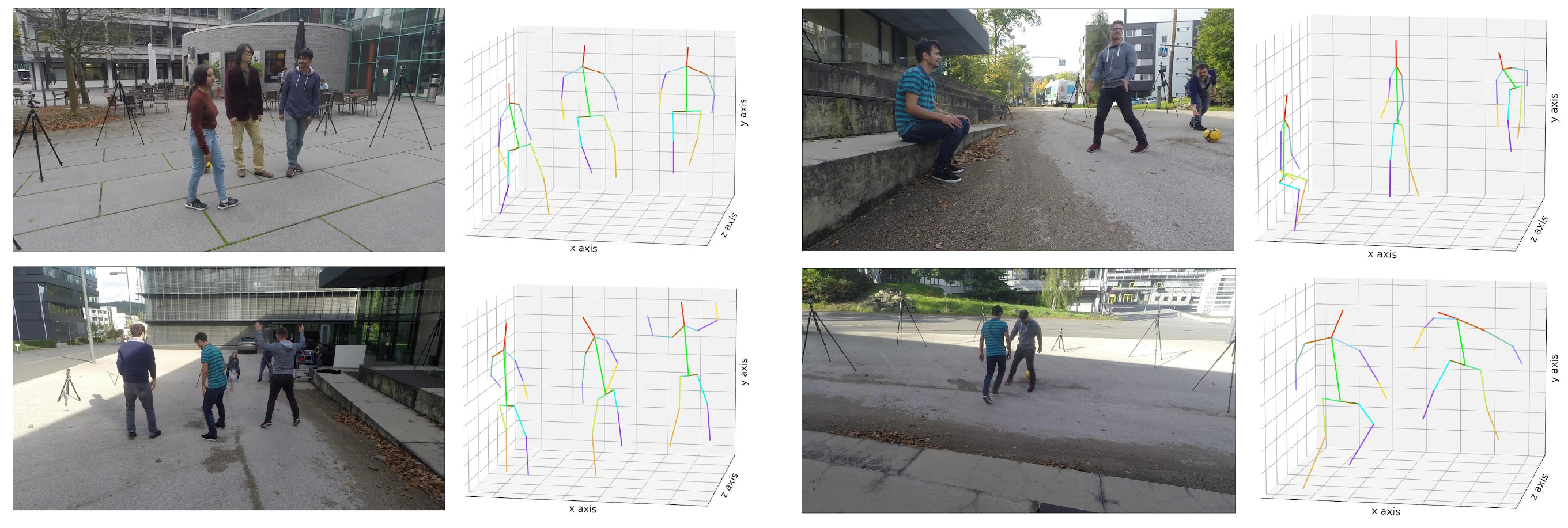}
\end{center}
\caption{ Qualitative results on MuPoTS-3D datasets. The test images and the corresponding 3D multi-person pose predicted by our proposed AdaptivePose-3D.}
\label{fig:3D_vis}
\end{figure*}

Furthermore, for the bottom-up methods, we outperform CMU-pose \cite{openpose} by a large margin. Compared with HigherHRNet \cite{higherhrnet} using the HRNet-W48 and higher output resolution, our methods achieve the equal performance only using small HRNet-W32 without multi-scale heatmap aggregation. Our method with HRNet-W48 improves HigherHRNet-W48 by 2.2 AP (1.6 AP) for single-scale (multi-scale) testing. Compared with the competitive DEKR \cite{dekr}, we achieve 1.2 AP gains without any bells and whistles (e.g., only using the center score as the final pose score) in inference stage. The results prove that the positioning capability of our network is much better.

\begin{table}
\begin{center}
{\caption{Comparisons with the state-of-the-art methods on the CrowdPose test set. $^{\dag}$ indicates multi-scale testing. }\label{tab:crowdpose}}

\resizebox{1.0\columnwidth}{!}{
\begin{tabular}{l|cccccc}
\hline
 Methods & $AP$ & $AP_{50}$ & $AP_{75}$ & $AP_{E}$ & $AP_{M}$ &  $AP_{H}$   \\
\hline
\multicolumn{7}{c}{Top-Down Methods}\\
\hline
Mask-RCNN~\cite{maskrcnn} &57.2& 83.5 &60.3& 69.4 &57.9& 45.8  \\
Rmpe~\cite{rmpe} & 61.0& 81.3& 66.0& 71.2& 61.4& 51.1 \\
SimpleBaseline~\cite{simplebaseline} & 60.8& 84.2& 71.5& 71.4& 61.2& 51.2\\
CrowdPose~\cite{crowdpose} &66.0& 84.2& 71.5& 75.5& 66.3& 57.4 \\ 

\hline
\multicolumn{7}{c}{Bottom-Up and Single-Stage Methods}\\
\hline
CMU-Pose~\cite{openpose} &- &-& -& 62.7& 48.7 &32.3 \\
HigherHRNet-W48~\cite{higherhrnet} &65.9& 86.4 &70.6 &73.3& 66.5& 57.9 \\
+ CenterGroup~\cite{centergroup} &67.6&87.7& 72.7&73.9&68.2&60.3\\
DEKR-W32~\cite{dekr} & 65.7& 85.7& 70.4& 73.0& 66.4& 57.5\\
DEKR-W48~\cite{dekr} &67.3& 86.4& 72.2 &74.6& 68.1 &58.7\\

\hline
Ours(DLA-34) &64.2&85.4&69.3&71.7&64.8&55.9 \\
Ours(HRNet-W32) &66.0&86.6&71.2&73.3&66.7&57.8 \\
Ours(HRNet-W48) &68.1&86.9&73.9&74.4&68.8&60.2 \\
\hline
\multicolumn{7}{c}{Bottom-Up and Single-Stage Methods$^{\dag}$}\\
\hline
HigherHRNet-W48$^{\dag}$~\cite{higherhrnet} &67.6& 87.4& 72.6& 75.8& 68.1& 58.9 \\

DEKR-W32$^{\dag}$~\cite{dekr} &67.0& 85.4& 72.4& 75.5 &68.0& 56.9  \\
DEKR-W48$^{\dag}$~\cite{dekr} &68.0 &85.5& 73.4& 76.6 &68.8 &58.4 \\
\hline
Ours(DLA-34)$^{\dag}$ &65.9&85.4&71.3&74.1&66.6&56.9\\
Ours(HRNet-W32)$^{\dag}$ &67.5&85.4&71.3&74.1&66.6&56.9 \\
Ours(HRNet-W48)$^{\dag}$ &69.2& 87.3 &75.0&76.7&70.0&60.9 \\
\hline
\end{tabular}

}
\end{center}
\end{table}

\subsection{AdaptivePose for 3D Pose Estimation}\label{subsection4.5}
We further extend AdaptivePose to 3D multi-person pose estimation \cite{SiE} to verify its generality.

{\bf Methodology.} To simply demonstrate the effectiveness of our proposed body representation and the single-stage network in 3D scenes. We use the pixel-wise depth map to predict the depth information of all human bodies. Based on the 2D AdaptivePose network, we further add two parallel branches. One is used to estimate a 1-channel root absolute-depth map in camera-centered coordinate system. For its target, the map values at the region centering on the root joint with radius 4 equal their absolute depths. The other branch is to output the 14-channel relative depth map of other keypoints compared to their root joint ( MuCo-3DHP and MuPoTS-3D only provide 15 keypoints annotations.). For its target, the map values at the region centering on the root joint with radius 4 equal their relative depths compared to root joints. Due to the visual perception of object scale and depth depends on the size of field of view (FoV), we normalize the depth by the size of FoV for all training samples as follows: Z = z * w /f , where Z is the normalized depth, z is the original depth, and f and w are the focal length and the image width.




During inference, first, we form the 2D human pose as described in subsection \ref{subsection3.3}. Second, we extract the absolute root depth and corresponding relative depth of the other keypoints via bilinear interpolation at root position of each pose candidates. The predicted depth values can be converted back to metric values during inference. Finally, according to the 2D keypoint coordinates and corresponding depth, the 3D pose can be reconstructed through the perspective camera model:
\begin{equation}
\small{\left[ X,Y,Z \right] ^T=ZK^{-1}\left[ x,y,1 \right] ^T,}
\end{equation}
where $\left[ X,Y,Z \right]$ refers to 3D coordinate in camera-centered coordinate system and $\left[ x,y \right]$ is 2D coordinate of a keypoint in pixel coordinate system, and $K$ is the camera intrinsic matrix.

{\bf Dataset.} MuCo-3DHP is the training dataset which is generated by compositing the 3D single-person pose estimation dataset MPI-INF-3DHP \cite{mpi}. MuPoTS-3D dataset is the test set contains 8,700 challenging images, which is generated out of doors and consists of 20 real-world scenes annotated with 3D keypoint positions. The annotations are obtained from a multi-view marker-less motion capture system.

{\bf Evaluation Metrics.} We leverage 3D percentage of correct keypoints (3D PCK$_{rel}$) with root alignment and the area under 3D PCK curve across different thresholds (AUC$_{rel}$) to evaluate relative root-centered prediction. The prediction is considered as correct if it lies within 15cm from the annotated keypoint position. We further use 3D PCK$_{abs}$ which indicates the 3D PCK without root alignment to evaluate the absolute camera-centered prediction. 

{\bf Implementation Details.} We adopt Adam optimizer to train our 3D network with the batch size 64 on a workstation with eight 32GB Tesla V100 GPUs. We employ warmup training strategy and the initial learning rate is set to 1.0e-3. The total training procedure is terminated at 20th epoch. Following previous work, we use MuCo-3DHP mixed with MS COCO to train the 3D network. All images are shuffled and each mini-batch are randomly sampled from the shuffled dataset. The input image are resized to 832$\times$512 for both training and testing process.

\begin{table}
\begin{center}
{\caption{Comparisons on the MuPoTS-3D~\cite{muco} dataset. All reported results are averaged over 20 test sequences. $\star$ indicates that the evaluations are conducted on all annotated persons.}\label{tab:3D}}

\resizebox{0.95\columnwidth}{!}{
\begin{tabular}{l|cccc}
\hline
 Methods & {PCK$_{rel} \uparrow $} & {PCK$_{abs} \uparrow$} & {AUC$_{rel} \uparrow$} & {PCK$_{rel}^{\star} \uparrow$}  \\
\hline
\multicolumn{5}{c}{Top-Down Methods}\\
\hline
3DMPPE~\cite{3dmppe} &82.5& 31.8 &40.9& {\bf 81.8}   \\
Hdnet~\cite{Hdnet} & 83.7& 35.2& -& - \\
Pandanet~\cite{Pandanet}  & -& -& -& 72.0 \\

\hline
\multicolumn{5}{c}{Bottom-Up Methods}\\
\hline

Xnect~\cite{XNect}  &75.8 &-& -& 70.4 \\

SMAP~\cite{smap}  &80.5& {\bf38.7}& 42.7 &73.5 \\

\hline
\multicolumn{5}{c}{Single-Stage Methods}\\
\hline
Ours(HRNet-W32) &{\bf 83.9} & 33.0 & {\bf 44.6} & 78.9 \\
\hline
\end{tabular}
}

\end{center}
\vspace{-4mm}
\end{table}

{\bf Results.} Table \ref{tab:3D} reports the results of our method and previous top-down and bottom-up methods on MuPoTS-3D \cite{muco} dataset. Our AdaptivePose-3D achieves 83.9 3D PCK$_{rel}$ and AUC$_{rel}$ score 44.6 with HRNet-W32, which outperforms top-down 3DMPPE \cite{3dmppe} and Hdnet \cite{Hdnet} by 1.4 and 0.2 3D PCK$_{rel}$ respectively. Compared with the bottom-up methods, we surpass Xnect \cite{XNect} and SMAP \cite{smap} by a large margin. The results prove that our body representation can more effectively build the relationship between human instance and corresponding keypoints than heuristic grouping process in 3D scene. Although we only use a particularly simple method to regress absolute root depth, we also achieve the promising performance on PCK$_{abs}$. We believe that our AdaptivePose-3D has great potential with more effective depth estimation approach.


\section{Conclusion}
In this paper, we introduce a fine-grained body representation that represents human parts as adaptive points. The novel representation can encode the various pose information more sufficiently and model the relations between human instance and keypoints more effectively. Based on the proposed body representation, we build a compact single-stage network named AdaptivePose. In inference stage, we eliminate the heuristic grouping as well as refinement process, and only need a single-step decode operation to form multi-person pose. Our network surpasses all existing bottom-up as well as single-stage approaches especially in case of without flip and multi-scale testing, and obtains the best speed-accuracy trade-offs on MS COCO. Comprehensive experiments prove the generality on crowd and 3D scenes.

\section*{Acknowledgments}
The paper is under the funding of National Nature Fund No.62071056, No.62102039 and No.62006244, and Young Elite Scientist Sponsorship Program of China Association for Science and Technology YESS20200140.

\vfill

\end{document}